\documentclass[conference]{IEEEtran}
\IEEEoverridecommandlockouts
\usepackage{cite}
\usepackage{amsmath,amssymb,amsfonts}
\usepackage{algorithmic}
\usepackage{graphicx}
\usepackage{textcomp}
\usepackage{verbatim}
\usepackage{comment}
\usepackage{epstopdf}
\usepackage{amsmath}
\usepackage{xcolor}
\usepackage{amssymb}

\usepackage{booktabs} 
\usepackage{subcaption}
\usepackage{hyperref}
\usepackage{multirow}
\usepackage{comment}
\usepackage{setspace}
\usepackage{xcolor}

\def\BibTeX{{\rm B\kern-.05em{\sc i\kern-.025em b}\kern-.08em
    T\kern-.1667em\lower.7ex\hbox{E}\kern-.125emX}}

\title{Performance Analysis and Characterization of Training Deep Learning Models on Mobile Devices}

\author{\IEEEauthorblockN{Jie Liu\IEEEauthorrefmark{0}, Jiawen Liu\IEEEauthorrefmark{0}, Wan Du\IEEEauthorrefmark{0} and Dong Li\IEEEauthorrefmark{0}}
\IEEEauthorblockA{\IEEEauthorrefmark{0}University of California, Merced\\\{jliu279, jliu265, wdu3, dli35\}@ucmerced.edu}}

\begin{document}

\maketitle

\begin{abstract}
Training deep learning models on mobile devices recently becomes possible, because of increasing computation power on mobile hardware and the advantages of enabling high user experiences. Most of the existing work on machine learning at mobile
devices is focused on the inference of deep learning models, but not training. The performance characterization of training deep learning models on mobile devices is largely unexplored, although understanding the performance characterization is critical for designing and implementing deep learning models on mobile devices.

In this paper, we perform a variety of experiments on a representative mobile device (the NVIDIA TX2) to study the performance of training deep learning models. We introduce a benchmark suite and a tool to study performance of training deep learning models on mobile devices, from the perspectives of memory consumption, hardware utilization, and power consumption. The tool can correlate performance results with fine-grained operations in deep learning models, providing capabilities to capture performance variance and problems at a fine
granularity. We reveal interesting performance problems and opportunities, including under-utilization of heterogeneous
hardware, large energy consumption of the memory, and high predictability of workload characterization. Based on the performance analysis, we suggest interesting research directions.

\end{abstract}

\section{Introduction}
\label{sec:intro}
Deep learning models have been widely deployed on mobile devices (e.g., mobile phones and smart home hub) to  process on-board sensing data and enable a variety of mobile applications (e.g., machine translation, speech recognition, cognitive assistance and street navigation)~\cite{lane2015deepear,mathur2017deepeye,xu2018deepcache}.
Those models are deployed for model inference (not for model training). The existing work has been conducted to analyze the performance and resource utilization of deep learning workloads on mobile devices when those models are deployed for model inference~\cite{lu2017modeling, hanhirova2018latency, adolf2016fathom, shi2016benchmarking, cnn-benchmarks, convnet-benchmark, Deepbench}. 
Those studies are important for optimizing the performance of deep learning models on mobile devices.  

Besides model inference, training deep learning models on mobile devices recently becomes possible, because of increasing computation power on mobile hardware and the advantages of enabling high user experiences. 
In particular, training deep learning models opens up a new approach to utilize the computational resources. As the hardware of mobile devices is increasingly powerful and domain-specific, especially with the emergence of artificial intelligence (AI) chipsets and powerful mobile GPU~\cite{Apple-A12, Snapdragon-855, Kirin-980, Xavier}, training deep learning models is moving from the cloud to mobile devices to leverage these decentralized computational resources. 
Furthermore, training deep learning models on mobile devices can avoid transmitting user data to the cloud as in the traditional method. The traditional method can cause the breach of user privacy (even using an anonymous dataset and mixing it with other data). 
For applications where the training objective is specified on the basis of data available on each mobile device, training on mobile devices can significantly reduce privacy and security risks by limiting the attack surface to only the device. 

Most of the existing work on machine learning at mobile devices is focused on the inference of deep learning models (particularly convolutional neural network (CNN) and recurrent neural network (RNN)), but not training. 
The performance characterization of training deep learning models on mobile devices is largely unexplored, although understanding the performance characterization is critical for designing and implementing deep learning models on mobile devices.

The recent work studies the performance of training deep learning models on servers~\cite{zhu2018benchmarking}.
However, training on mobile devices and on servers have different requirements, and have to be studied separately. First, training deep learning networks should not interfere with the user's regular operations on mobile devices; The interference can manifest as unexpected shorter battery life because of large energy consumption caused by training deep learning networks, or the extended latency of the user's operations. 
Second, the training data is likely to be collected and used for training on a daily basis. For example, to train a deep learning network for image classification, the system can use images collected per day as training samples and train the model every day. Third, a mobile device usually has a small memory capacity (compared with servers), hence some large deep learning models with tens of GB memory footprint (e.g., ResNet201 and VGG19) are not suitable to be trained on mobile devices.

In this paper, we perform a variety of experiments on a representative mobile device (the NVIDIA TX2) to study the performance of training deep learning models. Our study provides insightful observations and reveals potential research opportunities. In particular, this paper aims to answer the following research questions.


First, is training a deep learning network on a mobile device even possible? The existing work on federated learning has shown preliminary success of training some machine learning models on mobile devices \cite{bonawitz2019towards, smith2017federated, yang2018applied, konevcny2016federated, zhu2018multi, zhao2018federated, jeong2018communication,qi2018enabling, du2018efficient, sahu2018convergence, wang2018adaptive}. Different from a typical deep learning model, those machine learning models are small in terms of memory consumption and model size. Training deep learning models is known to be compute-intensive and memory-intensive.  Traditionally deep learning models are trained on servers with GPU with thousands of cores and high memory bandwidth. However, mobile devices are under recourse constraint, e.g., limited computation power and relatively small memory capacity. It is unknown whether and which deep learning models are trainable.


Second, how does training various deep learning models in mobile devices differ? Deep learning models have shown success in a broad range of application domains. Many deep learning models, such as DenseNet, Inception, ResNet, SqueezeNet and XceptionNet are related to image classification, which is one of the most common application domains for deep learning models. Other kinds of deep learning models, such as reinforcement learning, Generative Adversarial Network (GAN) that are used in other application domains such as robot controls, image generation and natural language processing, should also be taken into consideration. We aim to explore a diverse set of deep learning models in our study.


Third, what are the major performance problems when we train deep learning models on mobile devices? Are those problems on mobile devices different from those on servers? Answering the two questions is useful to identify research problems and train deep learning models more efficiently on mobile devices. 


By conducting extensive experiments with various deep learning models on a specific mobile device, the NVIDIA TX2, we find many insightful observations. In summary, we make the following contributions.

\begin{itemize}
    \item We introduce a benchmark suite for studying the workload of training deep learning models on mobile devices. The benchmark suite includes four application domains and includes ten common deep learning models. Those benchmarks are chosen with the consideration of possible resource constrained on mobile devices. We make our benchmark suite open-source and intend to continually expand it to support various mobile devices.
    
    \item We introduce a tool to study performance of training deep learning models on mobile devices, from the perspectives of the memory consumption, hardware utilization, and power consumption. More importantly, the tool can correlate performance results with fine-grained operations in deep learning models, providing capabilities to capture performance variance and problems at a fine granularity. 
    
    \item We reveal interesting performance problems and opportunities, including under-utilization of heterogeneous hardware, large energy consumption of memory, and high predictability of workload characterization. Based on the performance analysis, we suggest interesting research directions. 
 
\end{itemize}




\section{Training Deep Learning Models on Mobile Devices}
\label{sec:bg}

Deep learning is a general-purpose method that can be used to learn and model complicated linear and non-linear relationships between input datasets and output. Many deep learning models can be represented as a directed acyclic graph where nodes of the graph are connected neurons. Embedded in the graph, there are a number of parameters (e.g., ``weights'' and ``bias''). Those neurons and parameters are organized as layers. The process of obtaining the optimal values of those parameters to reach high prediction accuracy is called ``training''. Training involves many iterations of computation (sometimes millions of iterations), in order to acquire the optimal parameters. Each iteration is a \textit{training step}, and consists of forward and backward passes. During the backward pass, a backpropagation algorithm is used to optimize the parameters. The backpropagation algorithm calculates the gradient of each parameter. Also, the number of parameters and the corresponding gradients are equal. The intermediate results, which are generated by each layer, are dominated by feature maps. The forward pass generates feature maps, and the backward pass uses them to update the parameters. Hence, feature maps need to be temporarily stored in the memory during the forward pass and before the backward pass can consume them. 

The modern machine learning frameworks, e.g., TensorFlow~\cite{abadi2016tensorflow} and PyTorch~\cite{pytorch}, employ a dataflow graph where the deep learning models training is modeled as a directed graph composed of a set of nodes (operations). By decomposing a deep learning model graph (discussed above) into a set of nodes or fine-grained operations (e.g., SoftMax, Sigmoid and MatMul), these frameworks greatly improve hardware utilization and system throughput. Training a deep learning model easily involves a large number of operations. In a single training step of training deep learning models, there can be tens of different operations. Each operation can be invoked hundreds of times, each of which is an operation instance. 



\section{Methods}
\label{sec:methods}


In this section, we introduce the metrics and tools we use to evaluate the performance of training deep learning models.

\begin{figure*}[tb]
  \centering
    \includegraphics[
    width=\linewidth, 
    ]{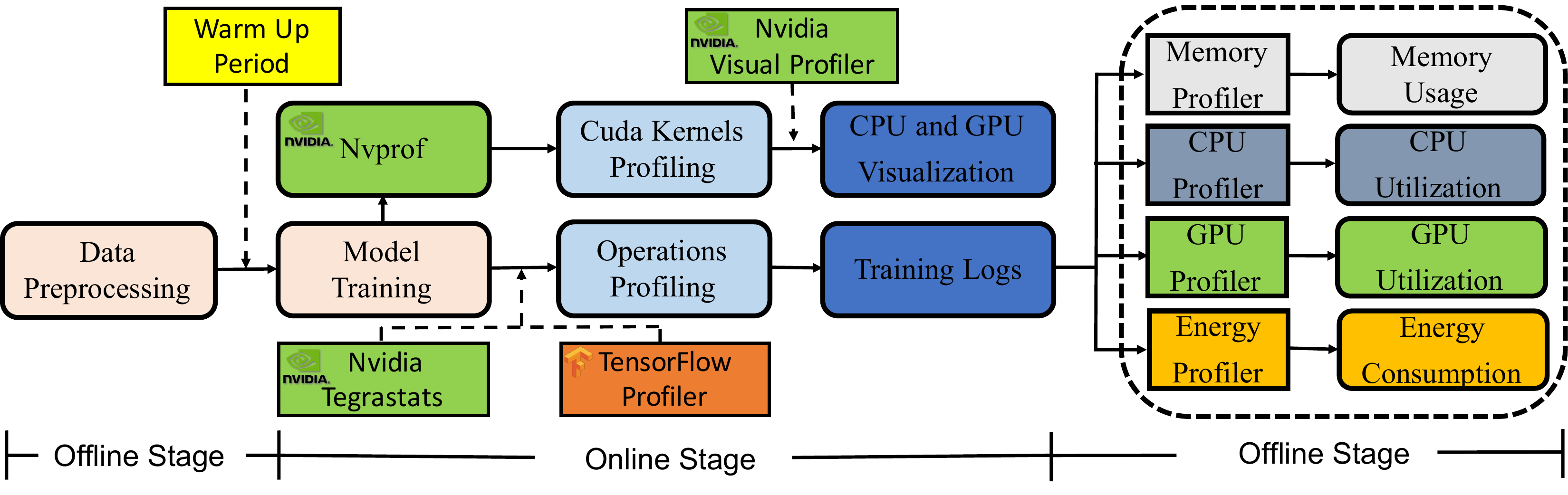}
        \caption{Profiling tools and profiling workflow.} 
  \label{fig:tool_chain}
\end{figure*}


\subsection{Evaluation Metrics}
\textbf{CPU utilization.} This metric quantifies how frequently CPU is utilized during deep learning models training. This metric is defined in Equations~\ref{eq:cpu_utilization_each_core} and~\ref{eq:cpu_utilization}.

\begin{equation}
\label{eq:cpu_utilization_each_core}
CPU\_Core\_Utilization = \frac{ T_{active}^{C} \times 100}{T_{total}}\%
\end{equation}

\begin{equation}
\label{eq:cpu_utilization}
CPU\_Avg\_Utilization = \frac{\sum_{i}^{n} ( CPU\_Core\_Utilization^{i})}{n}
\end{equation}

Equation~\ref{eq:cpu_utilization_each_core} is used to calculate the utilization of an individual CPU core. In Equation~\ref{eq:cpu_utilization_each_core}, $T_{total}$ denotes the total training time; $T_{active}^{C}$ indicates the active time of the CPU core.
Equation~\ref{eq:cpu_utilization} is used to calculate the average CPU utilization of all CPU cores. In Equation~\ref{eq:cpu_utilization}, 
$n$ is the total number of CPU cores for training deep learning models, and ${i}$ is the index of the CPU core.
A larger value of $CPU\_Avg\_Utilization$ indicates higher CPU utilization. We want high CPU utilization to achieve high throughout processing of training operations. 


\textbf{GPU utilization.}
This metric quantifies how frequently GPU is utilized during deep learning model training. This metric is defined in Equation~\ref{eq:gpu_utilization}.

\begin{equation}
\label{eq:gpu_utilization}
GPU\_Utilization = \frac{T_{active}^{G} \times 100}{T_{total}}\%
\end{equation}


As shown in Equation~\ref{eq:gpu_utilization}, the GPU utilization is defined similar to the CPU utilization in Equation~\ref{eq:cpu_utilization_each_core}. We also want high GPU utilization to achieve high throughout processing of training operations.


\textbf{Peak memory consumption.} Training deep learning models can be memory-consuming, as a large amount of parameters, derivatives and temporary variables use the memory space. Some popular deep learning models involve a large number of parameters. For example, VGG-16 and Resnet-50 have 138 million and 25 million parameters respectively, consuming 6.3 GB and 5.8 GB memory (the batch size is 64); SqueezeNet, a small deep learning model designed for mobile devices has 5 million parameters, consuming 5.7 GB memory (the batch size is 64). The memory consumption of a deep learning model sets up a constraint on whether training the model on a mobile device is feasible. The peak memory consumption is defined as the maximum memory usage during the training process. 


\textbf{Energy consumption.} 
Since a mobile device has limited battery life, reporting energy consumption of training deep learning models is critical to determine if the training is feasible within the battery life. Energy consumption is calculated based on Equation~\ref{eq:energy}. During the model training, we collect power consumption of the mobile device periodically. In Equation~\ref{eq:energy}, $time\_interval$ defines how frequently we collect power data, and $Power\_Consumption_{i}$ is the whole system power of the mobile device collected in a power sample data $i$.  

\begin{equation}
\label{eq:energy}
Energy = \sum_{i} time\_interval \times Power\_Consumption_{i}
\end{equation}

\textbf{Throughput.} This metric is used to evaluate the efficiency of the training process. Throughput in this paper is defined as how many training samples can be processed and used for training in one second. For example, when we train DenseNet40 using the batch size of 4, we can finish five training steps in one second, and each training step processes four images (samples). Hence, the throughput for training DenseNet40 is 20 samples per second.


\textbf{Idle state ratio for a core.} During the training, some CPU cores can be idle (i.e., the utilization is 0). Idle state ratio for a core is the percentage of the total training time that the core is idle. 



\subsection{Profiling Tools}
\label{sec:profiling_tools}
We use the existing tools, Nvprof~\cite{nvprof},  Tegrastats~\cite{tegrastats} and TensorFlow Profiler~\cite{tf-profiler}, for performance analysis and characterization of training deep learning models on the NVIDIA TX2. Nvprof is a profiling tool that collects execution time of computation on CPU and GPU. Nvprof can be used to identify idle or busy states of CPU and GPU. When used for GPU, Nvprof can also be used to identify GPU kernel names (hence the names of operations running on GPU). 

Tegrastats is a tool that collects hardware utilization of CPU and GPU, power consumption of hardware components (CPU, GPU, and memory) and memory consumption. 

TensorFlow Profiler ~\cite{tf-profiler} is a tool integrated into TensorFlow runtime to perform operations statistics, including operations execution time and dependency between operations. 

Nvprof and Tegrastats do not provide APIs that allow the programmer to trigger or stop profiling within the application. Nvprof and Tegrastats can run continuously as a system daemon and collect system-wide information at any moment. 
Simply using Nvprof and Tegrastats cannot meet the user's needs, because sometimes the user wants to correlate the profiling results (energy and memory consumption) with operations during the training process. The training process for deep learning models easily involves a large number of operations (thousands or even millions of operations in a single time step). Collecting the profiling results for operations is challenging.

We develop a tool to address the above challenge. In particular, during the training process, we record the start and end times of all operations at each layer; We also periodically examine the execution information (including CPU and GPU utilization, power consumption of hardware components) every 10ms (we choose 10ms to control the profiling overhead). The execution information is dumped into a file with the implicit timestamp information, due to our periodical profiling method. After the training process, we associate the execution information with operations based on timing information (i.e., the start and end times of all operations). Figure~\ref{fig:tool_chain} shows our tools and profiling workflow.

In Section~\ref{sec:results} (the section of evaluation results), the results 
are presented for all operations as a whole, because that allows us to easily present the results.




\subsection{Training Deep Learning Models on NVIDIA TX2}
\label{sec:training_on_TX2}
We use the NVIDIA TX2 as our evaluation platform. This platform is an embedded system-on-module (SoM) with a dual-core NVIDIA Denver2 plus a quad-core ARM Cortex-A57 (six CPU cores in total), eight GB LPDDR4 and integrated 256-core Pascal GPU (mobile GPU).  The GPU has two streaming multiprocessors (SM), and each SM has 128 cores. The eight GB memory is shared between CPU and GPU. The peak power consumption of TX2 is just 15 Watt. TX2 is a representative mobile platform. It has been commonly used in self-driving cars, robotics and drones. Many other common mobile devices, such as Snapdragon 855, Movidius Myriad2 and Nvidia Xavier, have comparable computation capability and memory capacity.
Table~\ref{tab:hardware-specification} summarizes the major hardware features of the NVIDIA TX2.

\section{Evaluation Results}
\label{sec:results}
In this section, we present the evaluation results and highlight major observations.

\subsection{Experiment Setup}

\begin{figure*}[htb]
  \centering
    \includegraphics[width=\linewidth, ]{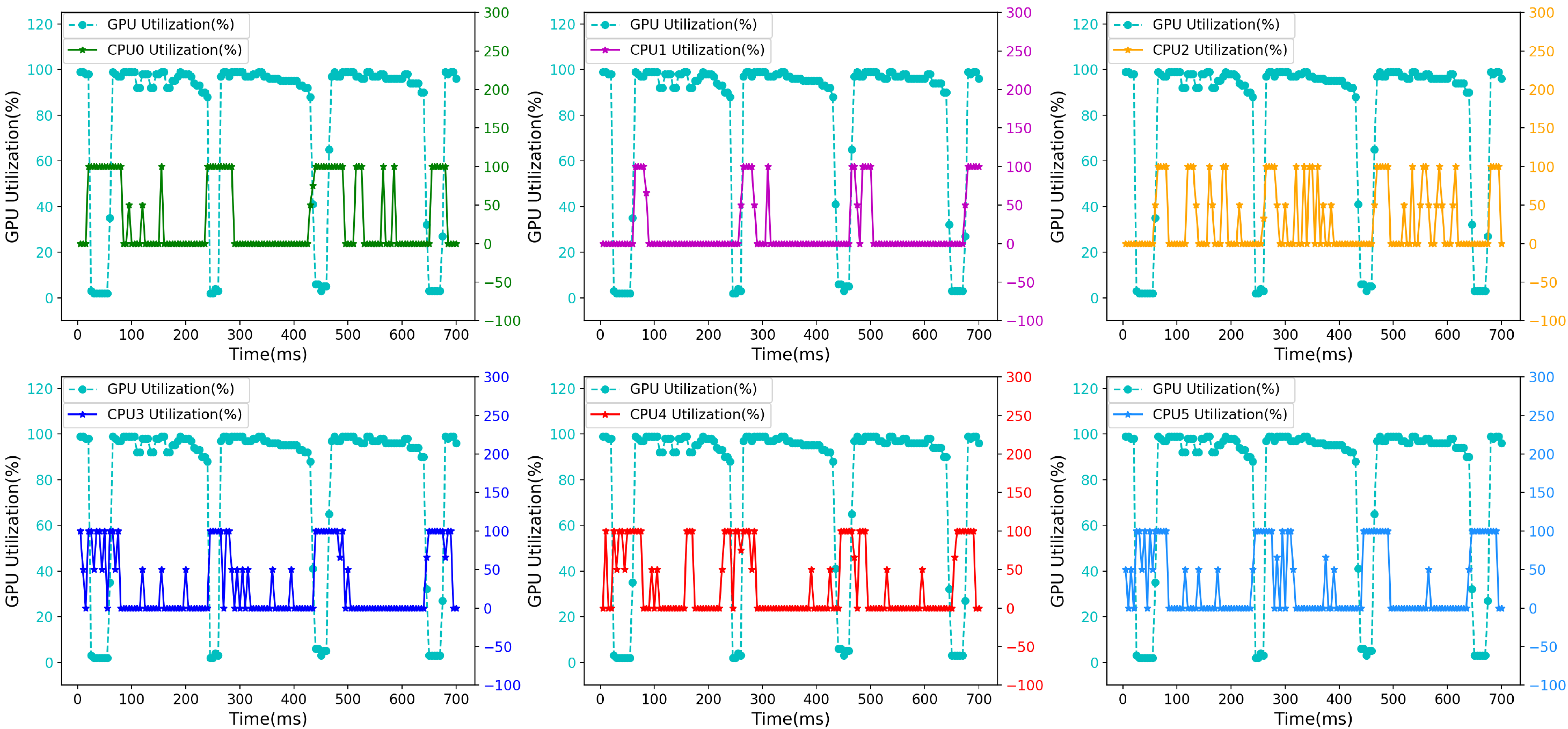}
        \caption{Utilization of GPU and six CPU cores for training Inception V1.} 
  \label{fig:CPU_GPU_inceptionV1}
\end{figure*}

\begin{figure*}[htb]
  \centering
    \includegraphics[width=\linewidth, ]{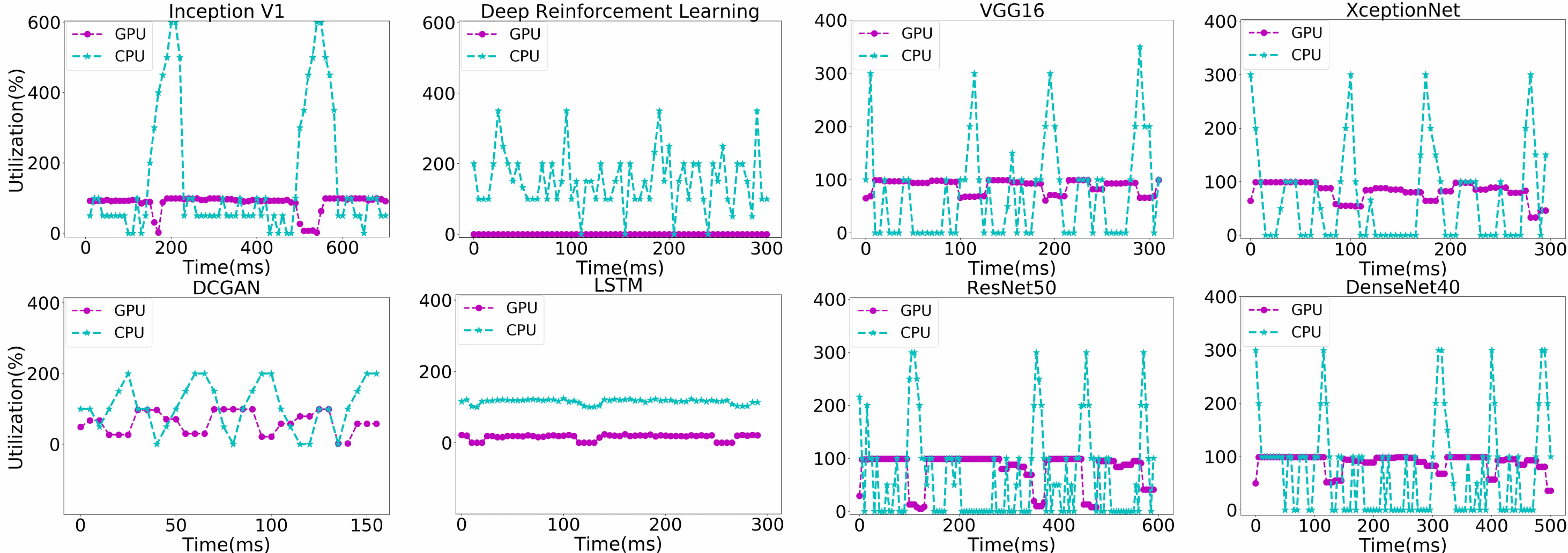}
        \caption{CPU and GPU utilization of different models.} 
  \label{fig:utilization_one_iteration}
\end{figure*}

Table~\ref{tab:hardware-specification} summarizes the deep learning models we use for evaluation.  The table also lists those deep learning models that cannot be successfully trained on TX2 because of the memory constraint. Among those models,  DenseNet100 and NMT can train for a few time steps, but have segmentation faults later on; VGG19, ResNet101, ResNet152 and BERT cannot get started on training at all.

 We use TensorFlow v1.13 to train the deep learning models. Unless indicated otherwise, we use the default configruations for TensorFlow. Note that we use TensorFlow instead of TensorFlow Lite, although TensorFlow Lite targets on mobile devices, because of the following reasons. (1) TensorFlow Lite only supports model inference, not training. Currently, there is no training framework especially targeting on training deep learning models on mobile devices. (2) TensorFlow and TensorFlow Lite have common implementations for many operations (e.g., convolution, matrix multiplication and max-pooling), especially those operations in the forward pass of some deep learning models. 

When reporting the performance, we skip the first three training steps, because they are often used by the TensorFlow runtime system to explore hardware architectures (e.g., cache capacities and memory access latency) for performance optimization. The performance of the first three training steps is not representative of other training steps.



\begin{table}
\small
\caption{The Specifications of NVIDIA TX2}
\label{tab:hardware-specification}
\resizebox{0.9\columnwidth}{!}
{

\begin{tabular}{|c|l|}
\hline
\textbf{Hardware}    & \textbf{Specifications}                     \\ \hline
Systems     & Tegra TX2 SoC                      \\ \hline
CPU1        & Quad-core ARM A57 MPCore           \\ \hline
Cache\_CPU1 & L1\_I: 128KB, L1\_D: 64KB, L2: 2MB \\ \hline
CPU2        & Dual-core NVIDIA Denver 2 64-Bit   \\ \hline
Cache\_CPU2 & L1\_I: 48KB, L1\_D: 32KB, L2: 2MB  \\ \hline
GPU         & NVIDIA Pascal GPU (256 CUDA Cores) \\ \hline
Memory      & 8GB 128-bit LPDDR4 Memory          \\ \hline
Storage     & 32GB eMMC 5.1                      \\ \hline
Power       & 7.5W / 15 W                        \\ \hline
\end{tabular}

}
\end{table}


\begin{table*}[t]
\caption{Descriptions for deep learning models in our evaluation}
\label{tab:hardware-specification}
\resizebox{1.0\textwidth}{!}
{

\begin{tabular}{|c|c|c|c|c|c|c|c|c|}
\hline
\textbf{Model}          & \textbf{\#Layers} & \textbf{Dominant layer} & \textbf{\#Flops} & \textbf{\#Parameters} & \textbf{Training dataset} & \textbf{Batch size} & \textbf{Successful training?} & \textbf{Application domain}          \\ \hline
DenseNet40\_12~\cite{huang2017densely} & 40       & CONV           & 30M     & 1M           & Cifar-10~\cite{krizhevsky2009learning}         & 1-64       &        \checkmark              & Computer Vision             \\ \hline
ResNet50~\cite{he2016deep}       & 50       & CONV           & 4G      & 98M          & Cifar-10         & 1-64       &            \checkmark          & Computer Vision             \\ \hline
SqueezeNet~\cite{iandola2016squeezenet}     & 40       & CONV           & 837M    & 5M           & Cifar-10         & 1-64       &       \checkmark               & Computer Vision             \\ \hline
VGG16~\cite{simonyan2014very}          & 16       & CONV           & 4G      & 134M         & Cifar-10         & 1-64       &          \checkmark            & Computer Vision             \\ \hline
XceptionNet~\cite{chollet2017xception}    & 39       & CONV           & N/A     & 23M          & Cifar-10         & 1-64       &     \checkmark                 & Computer Vision             \\ \hline
InceptionV1~\cite{szegedy2015going}     & 22       & CONV           & 30M     & 5M           & Cifar-10         & 1-64       &     \checkmark                 & Computer Vision             \\ \hline
Char-CNN~\cite{zhang2015character}       & 2        & LSTM+CONV      & 23M     & 1M           & Shakespeare~\cite{DBLP:conf/aistats/2017}      & 1-64       &           \checkmark           & Natural Language Processing \\ \hline
DCGAN~\cite{radford2015unsupervised}          & 40       & CONV           & 30M     & 1M           & Cifar-10         & 1-64       &     \checkmark                 & Image Generation            \\ \hline
Deep RL~\cite{mnih2015human}        & 4        & CONV           & N/A     & N/A          & Atari 2600 games~\cite{bellemare2012investigating} & 1-64       &         \checkmark             & Robotics Control            \\ \hline
AlexNet~\cite{krizhevsky2012imagenet}        & 8        & CONV           & 727M    & 60M          & Cifar-10         & 1-64       &     \checkmark                 & Computer Vision             \\ \hline
VGG19~\cite{simonyan2014very}          & 19       & CONV           & 20G     & 548M         & Cifar-10         & 1-64       &           x           & Computer Vision             \\ \hline
BERT~\cite{devlin2018bert}           & 12       & Embedding      & N/A     & 110M         & SQuAD~\cite{rajpurkar2016squad}            & 1-64       &            x          & Natural Language Processing \\ \hline
ResNet101~\cite{he2016deep}      & 101      & CONV           & 8G      & 155M         & Cifar-10         & 1-64       &             x         & Computer Vision             \\ \hline
ResNet152~\cite{he2016deep}       & 152      & CONV           & 11G     & 220M         & Cifar-10         & 1-64       &           x           & Computer Vision             \\ \hline
DenseNet100~\cite{huang2017densely}    & 100      & CONV           & 31M     & 7M           & Cifar-10         & 1-64       &          x            & Computer Vision             \\ \hline
seq2seq~\cite{sutskever2014sequence}        & 2        & LSTM           & 28G     & 348M         & IWSLT15~\cite{IWSLT15}          & 1-64       &            x          & Natural Language Processing \\ \hline
\end{tabular}

}
\end{table*}

\subsection{Performance Analysis}
We study the training performance from the following perspectives: hardware (CPU and GPU) utilization, power consumption, and peak memory consumption. 

\textbf{Hardware Utilization}

Figure~\ref{fig:utilization_one_iteration} shows the CPU and GPU utilization when we train the deep learning model Inception V1. The figure shows the hardware utilization for three training steps. Since the NVIDIA TX2 includes 6 CPU cores, we use six subgraphs to show the utilization of each of the six cores: the first two subgraphs show the utilization of two Denver2 cores, and the rest of them shows the utilization of four A57 cores. We have the following observations. 


\textit{Observation 1: The GPU utilization is generally much higher than the CPU utilization.}
In most of the times, the GPU utilization is close to 100\%, 
while each CPU core utilization ranges from 0\% to 60\%. Also, when the GPU utilization is high, the CPU utilization tends to be low, and vice versa. This indicates that the workload is not balanced well between CPU and GPU. There seems a lack of effective coordination between CPU and GPU. This observation is general and exists in many deep learning models (e.g., Inception V1, DCGAN, Resnet50, Xception).

Our further investigation reveals that when the GPU utilization is low, CPU is either busy with data fetching from storage (SSD) to main memory, or working on small operations that are not worth to be offloaded to GPU due to the large data copy overhead; When the GPU utilization is high, CPU is working on a few small operations, and most of CPU cycles are wasted.

Such an observation also exists in servers, but the difference is that the utilization difference between CPU and GPU on servers tends to be larger ~\cite{wu2019machine}, because GPU on servers are much more powerful than CPU on servers and hence more operations (after kernel fusing) tend to be scheduled on GPU.


\textit{Observation 2: The utilization of GPU and each core in CPU is predictable.} The utilization shows a periodical pattern where busy cycles alternate with less busy cycles. A period of the pattern corresponds to one time step. Across time steps, such a pattern repeatedly appears. This indicates that the computation across time steps remains stable and hence is highly predictable. This observation is consistent with the existing work that leverages predictability of deep learning workloads for performance optimization~\cite{Sivathanu:2019:AEP:3297858.3304072, ipdps19:liu}.

Such an observation also exists in servers. Since this observation is determined by the process of training deep learning models that repeatedly goes through a computation graph (and not hardware architecture-related), this observation is general and independent of hardware architectures.


 \begin{figure*}[htb]
  \centering
    \includegraphics[width=\linewidth, ]{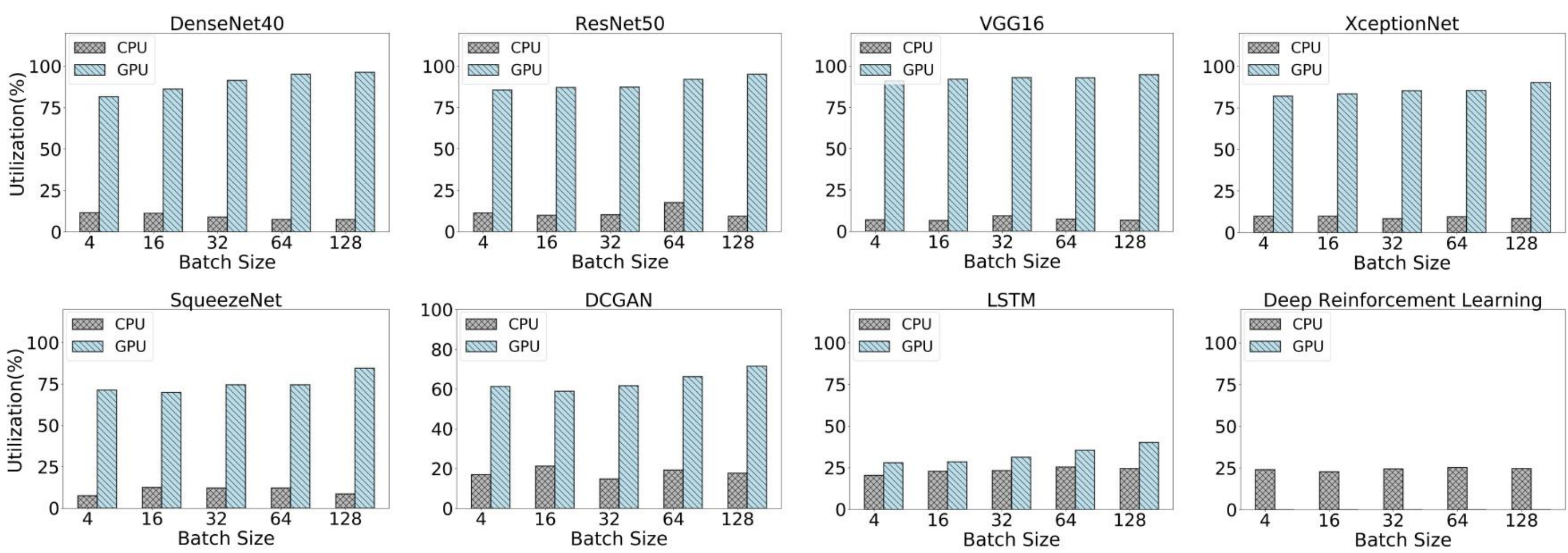}
        \caption{CPU and GPU utilization of different models.} 
  \label{fig:cpu_gpu_histogram}
\end{figure*}

\begin{figure}[htb]
  \centering
    \includegraphics[
    width=\linewidth, 
    height=.3\textheight
    ]{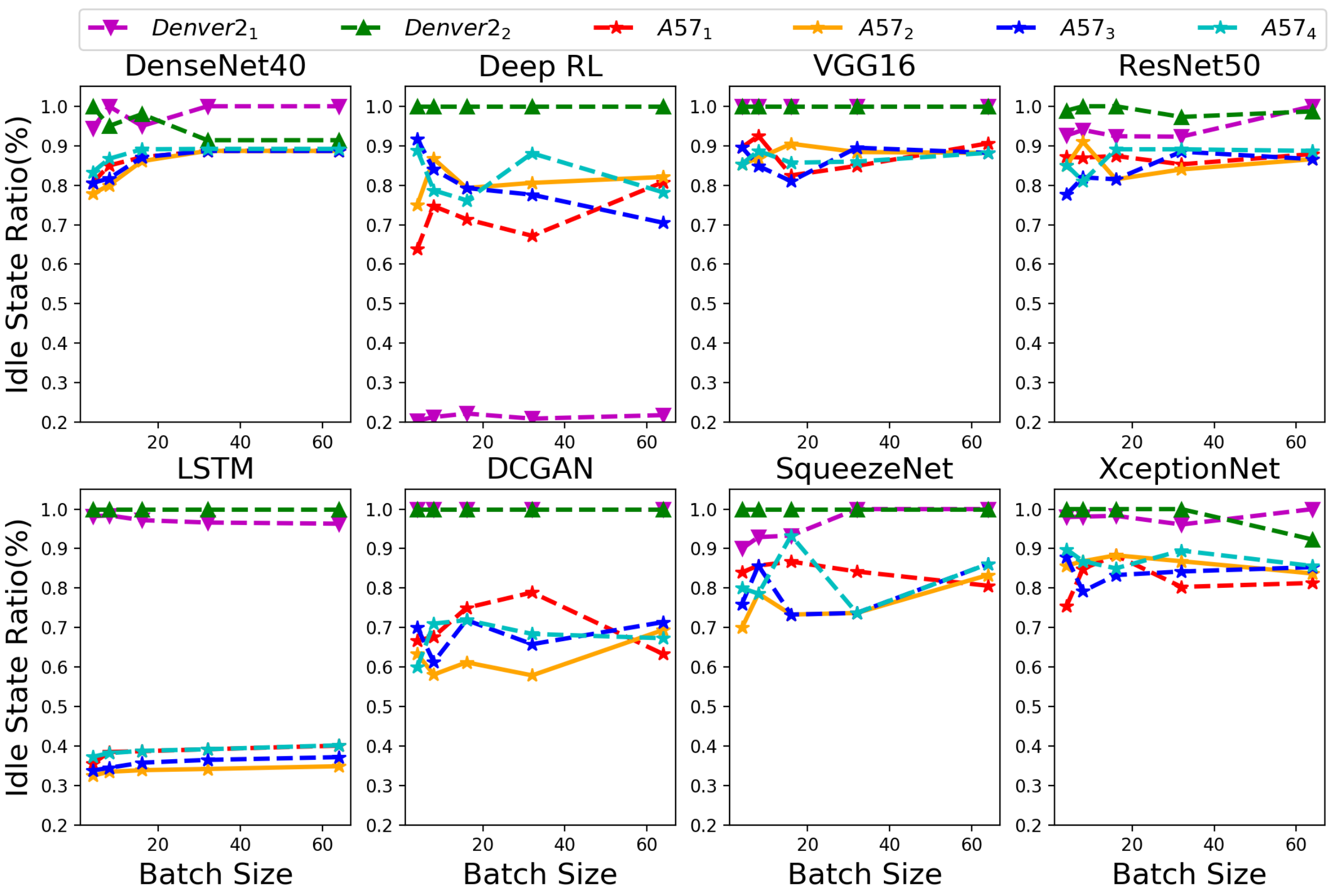}
        \caption{Idle state ratio of six CPU cores for different models.} 
  \label{fig:CPU_idle_test}
\end{figure}

 \begin{figure*}[htb]
  \centering
    \includegraphics[width=\linewidth, ]{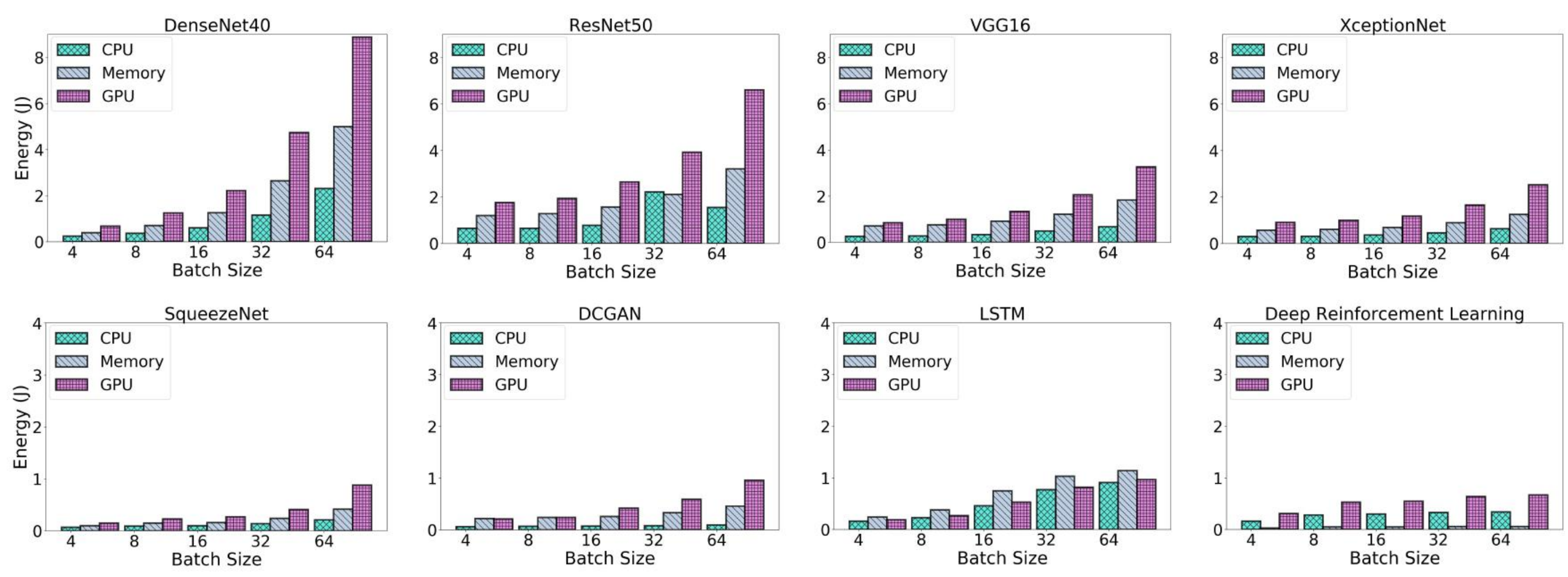}
        \caption{Energy consumption of different models.} 
  \label{fig:energy_consumption}
\end{figure*}

\begin{figure*}[htb]
  \centering
    \includegraphics[width=\linewidth, ]{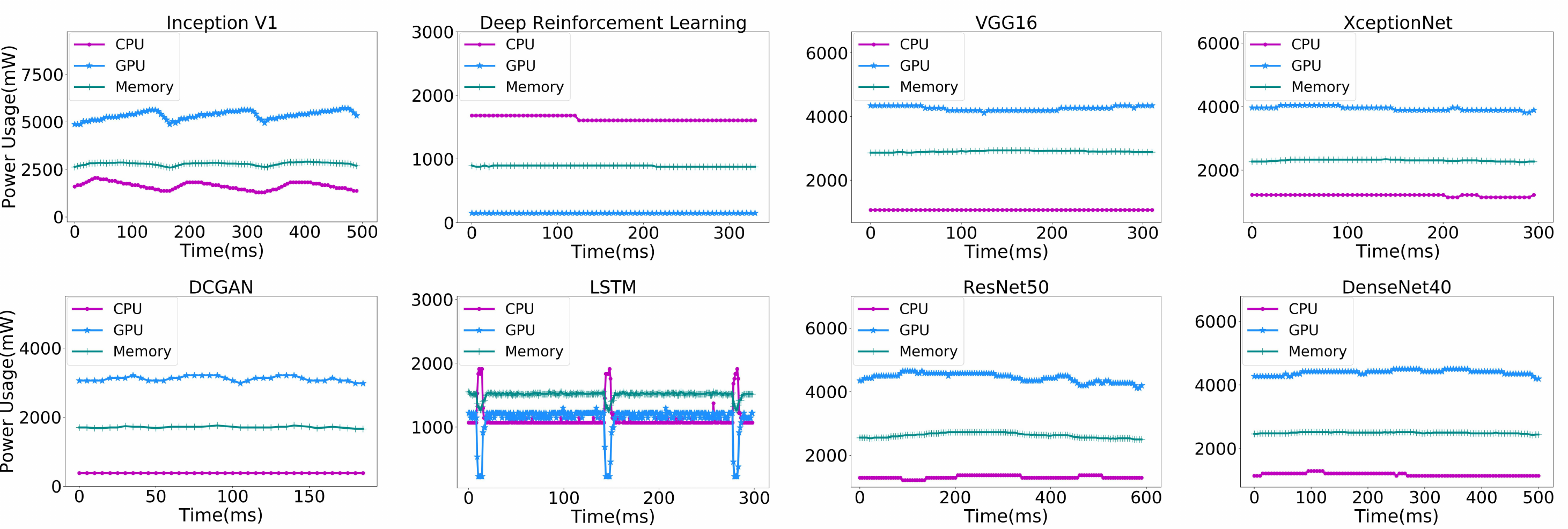}
        \caption{Power usage of different models in three iterations.} 
  \label{fig:power_iteration}
\end{figure*}

\textit{Observation 3: The GPU utilization is sensitive to the batch size, while the CPU utilization is not.} Figure  \ref{fig:cpu_gpu_histogram} shows the CPU and GPU utilization when the batch size changes. For DenseNet, the GPU utilization increases from $81.6\%$ to $96.4\%$ as the batch size changes from 4 to 64. For SqueezeNet and ResNet50, the GPU utilization increase from 71.4\% to 84.5\% and from 85.6\% to 95.6\% respectively, as we increase the batch size. However, for the CPU utilization, there is only 2.1\% difference on average across models.

As we increase the batch size, the memory footprint increases and computation for operations also increases. Since GPU works on most computation-intensive operations during the training, its utilization also increases as more computation requires more thread-level parallelism. The CPU utilization does not increase very much, because CPU works on small operations and most of data objects in those operations can be in caches. Slight increase of memory footprint due to the increase of the batch size does not cause extra cache misses and dos not significantly impact execution time.

Such an observation also exists in servers. But the variance of GPU utilization on servers does not change as much as that on mobile devices, because GPU on servers have more cores and hence offers more thread-level parallelism to work on increased computation as we increase the batch size.~\cite{ipdps19:liu}


\textit{Observation 4: Different cores have different utilization during the training.} TX2 has six heterogeneous cores: Two of them are Denver2 and four of them are A57. As Figure \ref{fig:CPU_idle_test} shows, the utilization of each core changes differently, as the batch size increases. There is no obvious correlation between the changes of the utilization across cores.  
We also notice that the two Denver2 cores have the highest idle state ratio (as high as 65\%) among all CPU cores, which indicates a large room for performance improvement.

We do not have the above observation on servers, because servers (especially x86 servers) usually do not have heterogeneous CPU cores~\cite{mittal2015survey}.


Figure\ref{fig:cpu_gpu_histogram} shows the GPU utilization of different deep learning models from different application domains. The models from the domain of computer vision have the similar GPU utilization, hence we show Inception V1 as a representative of this domain. We choose other models including LSTM and DCGAN to represent different application domains. 


\textit{Observation 5: The GPU utilization varies on different application domains.} 
In Figure \ref{fig:cpu_gpu_histogram}, we find the LSTM model (the domain of natural language processing) has a low GPU utilization (only about $25\%$), while the models from the domain of computer vision (e.g., ResNet) have higher GPU utilization (95\%).  Those models from computer vision has high GPU utilization, because they often employ convolution which is easy to leverage SIMT (Single Instruction Multiple Thread) on GPU and reach high GPU utilization. In LSTM, operations often have dependency and there is lack of available thread-level parallelism. 
The observation is consistent with the existing work\cite{zhang2018deepcpu, liu2018processing, jia2018beyond} that LSTM has lower utilization than computer vision models.

Such an observation also exists in servers. Since the GPU utilization is heavily impacted by the application domain, Observation 5 is general and independent of hardware architectures~\cite{zhu2018benchmarking}.


\textbf{Power and Energy Consumption}

Figures \ref{fig:power_iteration} and \ref{fig:energy_consumption} show power and energy consumption of GPU, CPU, and memory. Figure \ref{fig:power_iteration} shows how power consumption changes for three training steps. Figure \ref{fig:energy_consumption} shows energy consumption for one training step, when the batch size changes. Energy consumption is calculated based on Equation~\ref{eq:energy} with the time interval of 5ms. 


\textit{Observation 6: GPU is a power-consuming hardware component, but for some deep learning model, the memory consumes more power than GPU.}
Figure \ref{fig:cpu_gpu_histogram} shows that for the domain of computer vision, GPU is the most time-consuming hardware component when we train deep learning models (especially CNN models) such as  ResNet50 and VGG16. In those models, GPU consumes $4\times$ and $2\times$ of power consumption of CPU and memory respectively. GPU consumption can take up to $57.4\%$ of the whole system power. Different from the above examples, the memory is the most power-consuming hardware component (not GPU), when we train LSTM. Compared with the CNN models, LSTM has relatively bad data locality and causes more intensive memory accesses. As a result, the memory, shared between GPU and CPU, draws large power consumption.

Such an observation does not exist in servers. In servers, GPU (including its global memory) is the most power-consuming (e.g., NVIDIA V100 takes up to 250 Watt (more than half of the system power) when training LSTM, while the memory (main memory) takes only 20\%-30\% of the total system power.)

\textit{Observation 7: The power consumption across training steps is predictable.} 
Similar to the hardware utilization, the power consumption of hardware components shows a periodical pattern. This pattern is highly predictable across training steps. Figure \ref{fig:power_iteration} shows such results. On servers, we have the similar observations.

\textit{Observation 8: As we increase the batch size, the energy consumption increases as well, but not in a proportional way.} Figure {\ref{fig:energy_consumption} shows the results to support this observation. As we increase the batch size from 4 to 64 (16x increase), the energy consumption of the whole system increases as well. However, the increase of the energy consumption is at least 2.2x and at most 10.5x, less than 16x when we change the batch size.

Also, different models show quite different energy consumption. Among the five deep learning models for computer vision, DenseNet40 is the most energy-consuming one, while the Squeezenet is the most energy efficient one. The above conclusion is true as we run the training to completion (including all time steps). The above observation also exists in servers.

Consistent with the results of power consumption, we notice that for some models (e.g., DenseNet40), GPU is the most energy-consuming one, while for LSMT, the memory is the most energy-consuming one. 


\textbf{Peak Memory Consumption}

The memory is one of the key limiters for deep learning models training on mobile devices. Some large models, e.g., ResNet101 and VGG19, consume more than 10 GB memory for training, while TX2 only has 8 GB memory. Those models cannot be trained on TX2. In our study, we aim to study the impact of the batch size on the memory consumption of deep learning models. Different from on servers, on mobile devices we must carefully choose the batch size, not only for good training accuracy as on servers, but also for acceptable memory consumption.

For training (especially CNN and RNN), the memory is consumed by the following critical variables: parameters (including weights and bias), gradients, input data, and intermediate data. Among them, the intermediate data is the most memory consuming. The intermediate data includes the work space and feature map. The work space is the memory consumed by the machine learning framework (e.g., TensorFlow or PyTorch). The memory consumption of the work space varies for different frameworks. The feature map, sitting in the middle of two neighbor layers of a CNN or RNN model, is generated by one layer, and used as input of the next layer.


\textit{Observation 9: Choosing a good batch size is critical to be able to train deep learning models on mobile devices.} Figure \ref{fig:memory_usage} shows memory usage as we change the batch size. As expected, parameters, input data and gradients remain constant, as we increase the batch size. But the memory consumption of intermediate data increases significantly, as we increase the batch size. For example, for DenseNet40, when the batch size increases from 4 to 64, the memory consumption of intermediate data increases from 2.2 GB to 5.9 GB. When we use larger batch sizes (12nd 256), we run out of memory for all models. 


\begin{figure*}[htb]
  \centering
    \includegraphics[width=\linewidth, ]{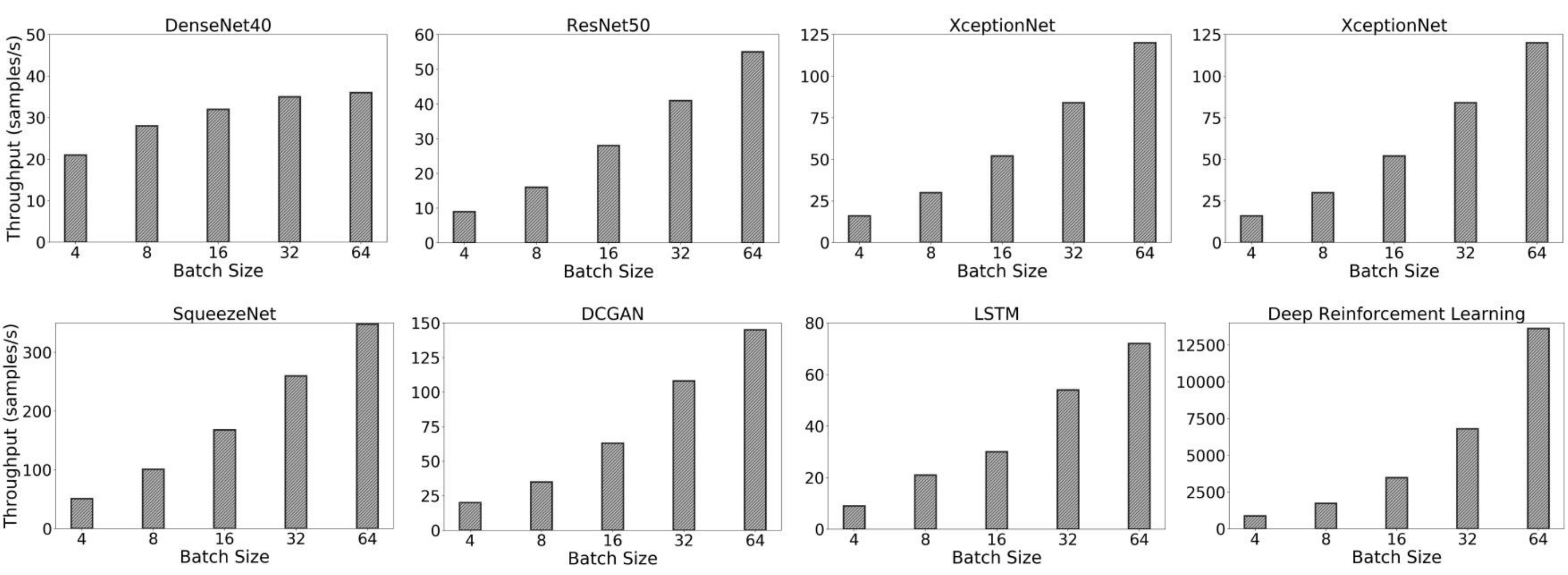}
        \caption{Throughput of deep learning models.} 
  \label{fig:throughput}
\end{figure*}

\begin{figure}[htb]
  \centering
    \includegraphics[width=\linewidth, ]{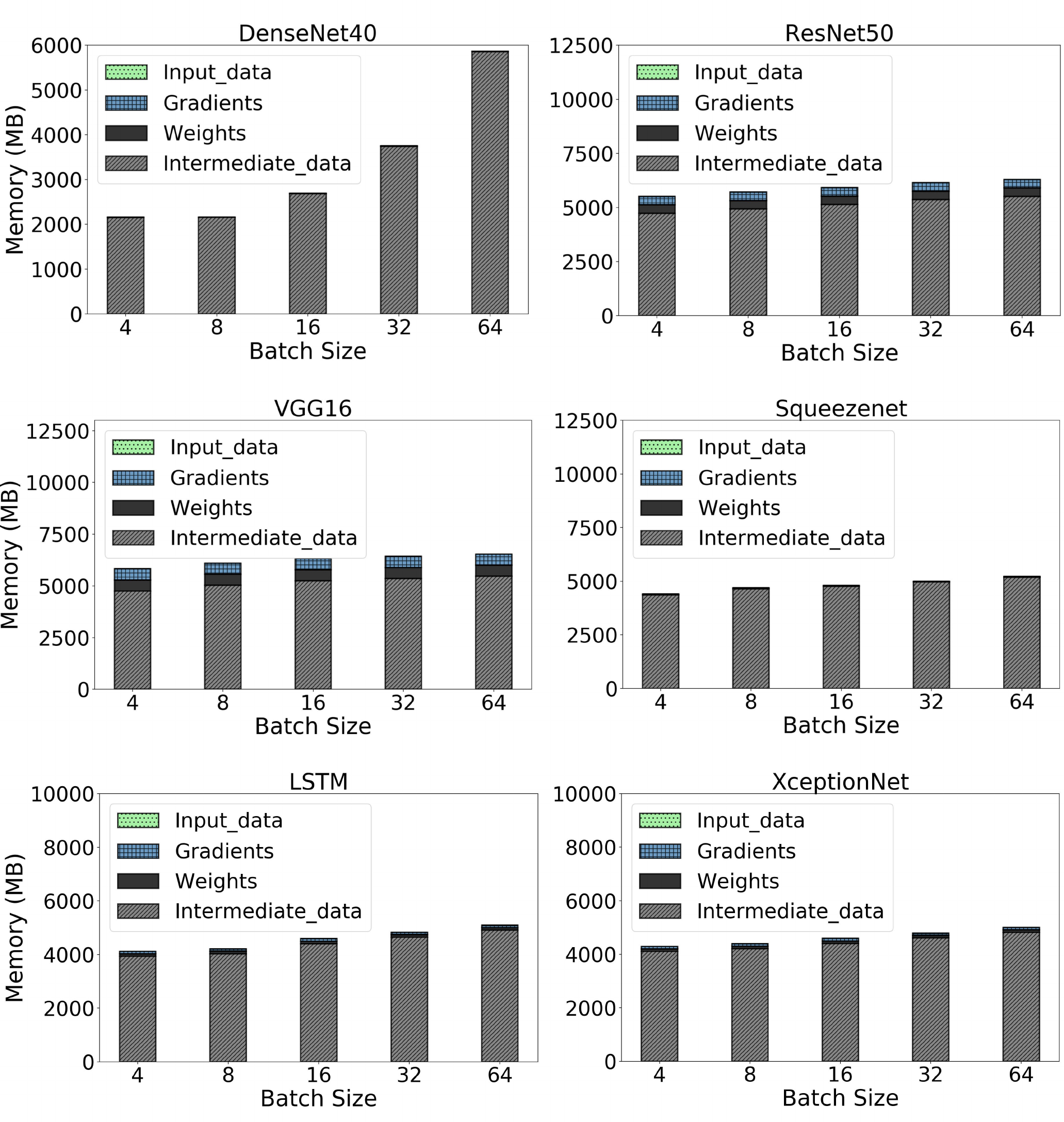}
        \caption{Memory usage of deep learning models.} 
  \label{fig:memory_usage}
\end{figure}

\begin{figure}[tb]
  \centering
    \includegraphics[width=\linewidth, ]{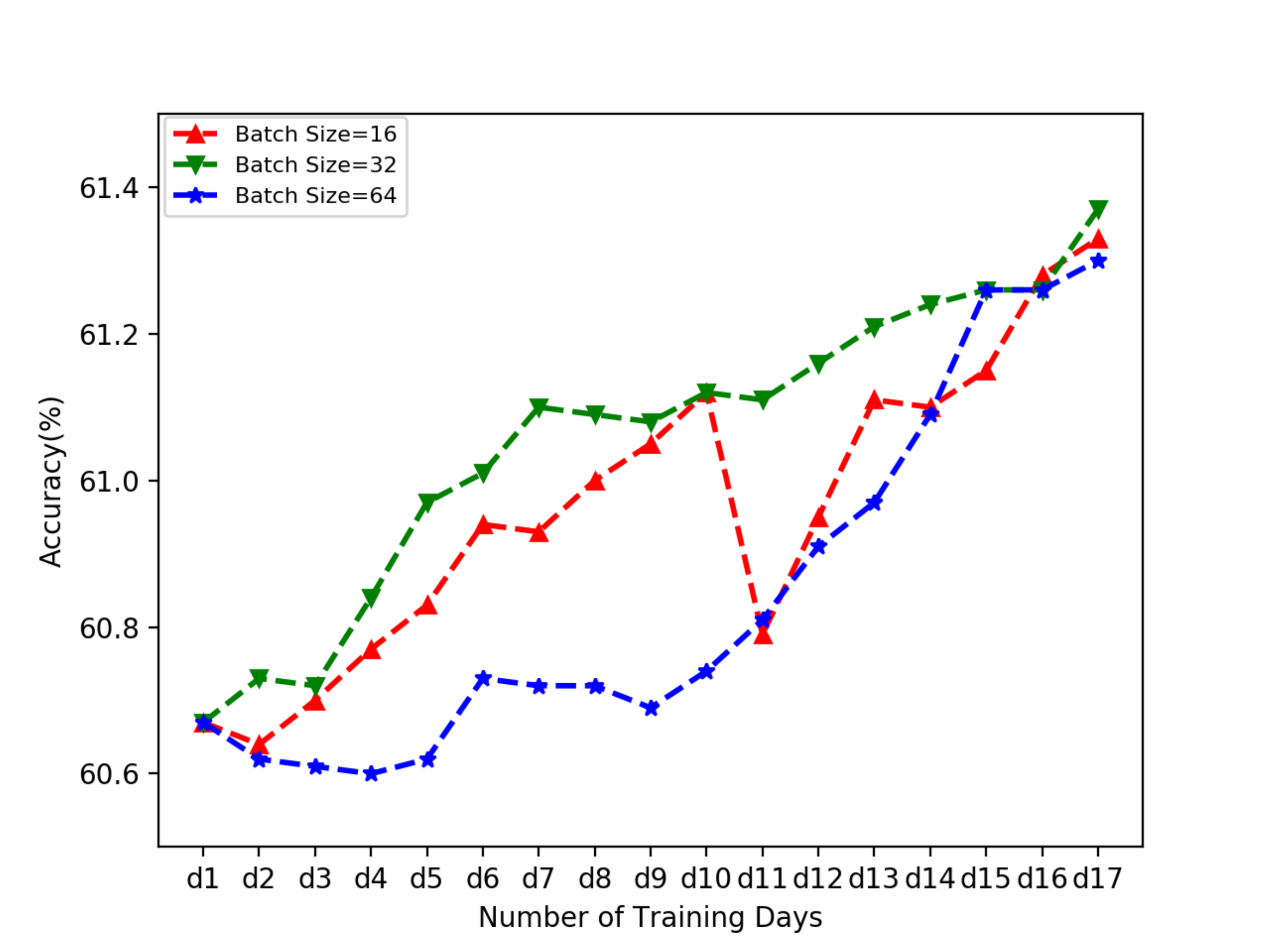}
       \vspace{-20pt}
        \caption{The accuracy variance as we increase training samples at a daily base to train DenseNet40.} 
  \vspace{-10pt}
  \label{fig:accuracy}
\end{figure}

\textbf{{Throughput}}

To quantify throughput, we employ a common metric, training samples per second, instead of using images per iteration (training step) as in some deep learning models, because of the following two reasons. First, our collection of deep learning models includes CNN, RNN and Deep reinforcement learning models, which means that the training samples for some models are not images. For example, the training samples are sentences for some RNNs (e.g., seq2seq).
Second, as we change the batch size for evaluation, the number of training samples for a training step changes as well, which indicates that the execution time per training step (iteration) changes. Hence, using ``second'' (the time metric) instead of ``iteration'' makes more sense.

\textit{Observation 10: Throughput increases as the batch size increases.} Figure~\ref{fig:throughput} shows the throughput as we change the batch size. 
For all models, the throughput increases as the batch size increases. For example, for ResNet50, the throughput increases from 9 to 55 samples per second as the batch size increases from 4 to 64.

The above observation can also be seen in servers~\cite{zhu2018benchmarking}.

\textit{Observation 11: Across models, throughput changes differently, as we increase the batch size.} In Figure~\ref{fig:throughput}, the throughput of the deep reinforcement learning model increases from 889 to 13,618 samples per second as the batch size increases from 4 to 64 (15.3x speedup). However, for DenseNet and ResNet50, such throughput speedup is 1.7x and 6.1x, respectively. The deep reinforcement learning model has big throughput speedup as we increase the batch size. This is because the training time of the deep reinforcement learning does not change too much, as we increase the batch size. As a result, the throughput increases significantly, as we increase the batch size. 

The above observation also exists on servers~\cite{zhu2018benchmarking}.

\textbf{Study on modeling accuracy}

Training a deep learning model on a mobile device is different from that on a server, because training samples can be dynamically generated when the mobile device is used. For example, training DenseNet40 can be based on training samples (images) collected at the user's day time. In this evaluation, we evaluate a scenario where the user uses a mobile device to generate 64 images per day, and those images are used to train a deep learning model (DenseNet40). We also assume that the model, before started to be trained, is already trained on a server, but needs to be trained further, using the user's new training samples. Figure~\ref{fig:accuracy} shows the variance of the accuracy, as we use the above training method for 17 days. As the day 0, the training accuracy is 60.65\%, because the model is already trained on a server. 

\textit{Observation 12: Training a deep learning model on a mobile device can slowly increase training accuracy.} Figure~\ref{fig:accuracy} reveals that the accuracy of DenseNet40 increases from 60.65\% to 61.32\% (using three different batch sizes). The above observation reveals that using the above method does slowly increases the accuracy. In this special scenario, depending on whether the user has high requirement on the model accuracy, the training on the mobile device can continue as more training samples are collected or stop.



\section{Discussion and Future Research Directions}

\textbf{Feasibility of training deep learning models on mobile devices.}
Our work demonstrates the feasibility of training some deep learning models on a mobile device. Most of the models we study are traditionally trained on a server, and seldom trained on any mobile device. Those deep learning models come from various application domains, and have potential to provide new services for mobile users. 

Our observation 9 reveals that choosing an appropriate batch size has a big impact on whether training a deep learning model is feasible. Furthermore, it is well known that the batch size has an impact on the model accuracy. Hence, there is a non-trivial tradeoff between the training feasibility and accuracy on mobile devices. Such a tradeoff deserves further study.

\textbf{Hardware utilization.}
Mobile devices often offer rich hardware heterogeneity (e.g., there are two types of CPU cores in the NIVIDA TX2), richer than x86 servers. However, such hardware heterogeneity is not leveraged well in the current machine learning frameworks. This fact is pronounced by two observations: (1) The utilization of all CPU cores is relatively low (comparing with GPU); (2) The heterogeneity of CPU cores is completely ignored. As a result of such fact, the CPU cycles are wasted and the computation power of specialized CPU cores (e.g., ARM Cortex-A57) is not fully utilized. 

The recent work from Facebook~\cite{wu2019machine} reveals that the performance difference between CPU and GPU on mobile devices is smaller than that on servers. Based on this work and our observations, we see great opportunities to improve the current scheduling mechanism in the machine learning frameworks. Only using GPU for computation-intensive operations may not be a good scheduling strategy. Instead, balancing workloads on CPU and GPU to maximize system throughput (for finishing operations) is a better one.

\textbf{Energy consumption.}
Mobile devices are more sensitive to energy consumption than servers. Training deep learning models on mobile devices raises concerns on whether the battery life is good enough to support training. Although the recent work suggests to train deep learning models when the mobile device is charged ~\cite{bonawitz2019towards, DBLP:journals/corr/abs-1902-00146, mcmahan2016communication}, the charging time can be longer than the training time. The batter may still be needed to finish training. Hence, minimizing energy consumption during training is critical. Our observation reveals that the memory can be more energy-consuming than CPU and GPU, when we train some deep learning networks. Reducing energy consumption of the memory is necessary for mobile devices. How to reduce energy consumption of the memory without impacting performance (execution time) is an open topic.  

\textbf{Predictability of workload characterization.}
The workload of training deep learning networks is predictable, which means execution time, hardware utilization and power consumption show a repetitive pattern across training steps. Such predictability allows us to apply dynamic profiling on a few training steps to collect workload characterization, based on which we guide operation scheduling and power management in the future training steps. Predictability of execution time during the training has been leveraged in the existing work~\cite{ipdps19:liu, Sivathanu:2019:AEP:3297858.3304072}. We expect to leverage the predictability of other characterization in the future work.

\section{Related Work}

\textbf{Performance optimization of deep learning model training.}
Some recent works \cite{konevcny2016federated, zhu2018multi, zhao2018federated, jeong2018communication,qi2018enabling, du2018efficient, sahu2018convergence, wang2018adaptive, bonawitz2019towards, smith2017federated, yang2018applied, mao2017adalearner} have demonstrated the promise of training neural networks (NN) on mobile devices. They are focused on exploring performance optimization in the perspectives of algorithm and system.  
For example, Mao et al. \cite{mao2017adalearner} implement a distributed mobile learning system that trains a neural network by multiple devices of the same local network in parallel. 
They design a scheduler to adapt the training configuration for heterogeneous mobile resources and network circumstances. 
Bonawitz et al. \cite{bonawitz2019towards} develop a federated learning system to achieve NNs training on mobile platforms using TensorFlow. Some practical issues have been addressed, e.g., local data distribution, unreliable device connectivity and limited on-board resources. 
Kone{\v{c}}n{\`y} et al. \cite{konevcny2016federated} use parameter compression techniques to reduce the uplink communication costs in federated learning. 
This paper is orthogonal to the above works. Our comprehensive model profiling and analysis can be used to develop more efficient NN training schemes on mobile devices. 

Zhu et al. \cite{zhu2018benchmarking} study the training performance and resource utilization of eight deep learning model models implemented on three machine learning frameworks running on servers (not mobile devices) across different hardware configurations. However, they do not consider power and energy efficiency.
In contrast, our work is focused on deep learning models training on mobile devices.

\textbf{Profiling of deep neural network inference.}
Many works have been conducted to analyze the performance and resource utilization of machine learning workloads (inference, not training) on mobile devices \cite{lu2017modeling, hanhirova2018latency, adolf2016fathom, shi2016benchmarking, cnn-benchmarks, convnet-benchmark, Deepbench}. 
Lu et al. \cite{lu2017modeling} measure the performance and resource usage of each layer in CNNs running on mobile CPUs and GPUs. 
Based on the results of profiling and modeling, they implement a modeling tool to estimate the compute time and resource usage of CNNs. However, they only consider CNNs, but not RNNs or reinforcement learning models which are also important for mobile applications.  
Hanhirova et al. \cite{hanhirova2018latency} profile  the performance of multiple CNN-based models for object recognition and detection on both embedded mobile processors and high-performance server processors. They find that there exists significant latency–throughput trade-offs. Unfortunately, the above works only study the inference of CNNs. On the contrary, we profile and analyze the performance and resource requirements of CNNs, RNNs and deep reinforcement learning models training on mobile devices.


\section{Conclusions}
\label{sec:conclusions}

Training deep learning networks on mobile devices is emerging because of increasing computation power on mobile hardware and the advantages of enabling high user experiences. The performance characterization of training deep learning models on mobile devices is largely unexplored, although understanding
the performance characterization is critical for designing and implementing deep learning models on mobile devices. This paper is the first work that comprehensively studies the performance of training deep learning network on a mobile device. Our study is based on a set of profiling tools on mobile devices, and uses a set of representative deep learning models from multiple application domains. We reveal many research opportunities as a result of our study. We hope that our study can motivate future study on optimizing performance of training deep learning networks on mobile devices.  

\bibliographystyle{ieeetr}
\bibliography{ref}

\end{document}